\documentclass[12pt]{article}
\usepackage{template}

\begin{document}

\thispagestyle{empty} 
\baselineskip=28pt

\begin{center}
    {\LARGE{\bf Variational Autoencoded Multivariate Spatial Fay-Herriot Models}}
\end{center}

\baselineskip=12pt

\vskip 2mm

\begin{center}
    Zhenhua Wang\footnote{(\baselineskip=10pt to whom correspondence should be
  addressed)  Department of Statistics, University of
  Missouri, 146 Middlebush Hall, Columbia, MO 65211,
  zhenhua.wang@missouri.edu}\ ,
Paul A. Parker\footnote{\baselineskip=10pt Department of Statistics, University of California, Santa Cruz, 1156 High Street,
  Santa Cruz, CA 95064, paulparker@ucsc.edu}\ , and 
Scott H. Holan\footnote{\baselineskip=10pt Department of Statistics, University of
  Missouri, 146 Middlebush Hall, Columbia, MO 65211,
  holans@missouri.edu}\,\footnote{\baselineskip=10pt Office of the Associate Director for Research and Methodology, U.S. Census Bureau, 4600 Silver
  Hill Road, Washington, D.C. 20233, scott.holan@census.gov}\end{center}

\vskip 4mm

\begin{center}
    {\bf Abstract}
\end{center}
Small area estimation models are essential for estimating population characteristics in regions with limited sample sizes, thereby supporting policy decisions, demographic studies, and resource allocation, among other use cases. The spatial Fay-Herriot model is one such approach that incorporates spatial dependence to improve estimation by borrowing strength from neighboring regions. However, this approach often requires substantial computational resources, limiting its scalability for high-dimensional datasets, especially when considering multiple (multivariate) responses. This paper proposes two methods that integrate the multivariate spatial Fay-Herriot model with spatial random effects, learned through variational autoencoders, to efficiently leverage spatial structure. Importantly, after training the variational autoencoder to represent spatial dependence for a given set of geographies, it may be used again in future modeling efforts, without the need for retraining. Additionally, the use of the variational autoencoder to represent spatial dependence results in extreme improvements in computational efficiency, even for massive datasets. We demonstrate the effectiveness of our approach using 5-year period estimates from the American Community Survey over all census tracts in California.

\baselineskip=12pt 

\vskip 4mm

\par\vfill\noindent

\clearpage
\pagenumbering{arabic}
\baselineskip=24pt

\section{Introduction}

Small area estimation focuses on estimating population characteristics for regions with limited sample sizes, where direct estimates may be unreliable due to sampling error. As one of the most widely used methods in small area estimation, the Fay-Herriot (FH) model assumes that the direct estimate is a noisy version of the true population quantity, where the noise has a known variance  equal to the sampling error variance. \citep{fay1979estimates}. However, the standard Fay-Herriot model can be still exhibit significant uncertainty in cases with very limited sample sizes, high sampling variances, or when there are not strong predictor covariates available. Recent literature shows that estimation accuracy can be enhanced by borrowing strength from neighboring regions, which is commonly known as the spatial FH model. \citet{salvati2004small} constructed the spatial FH model using a simultaneous autoregressive structure. \citet{porter2014spatial, porter2015small} incorporate spatial structure in the FH model using a conditional autoregressive (CAR) prior. \citet{chandra2015spatially} developed the empirical best linear unbiased predictor for a nonstationary spatial FH model.
However, a key limitation is that sampling from the posterior distribution of the FH model with a spatial prior requires substantial computational resources and excessive run times, which makes it impractical for scaling up to high-dimensional geographic datasets.

This computational issue becomes even more pronounced when handling multiple responses. The most straightforward approach for handling multiple responses in a CAR model is the separable multivariate CAR (MCAR) model, where the spatial covariance matrix is ``separated" from the covariance among responses \citep{gelfand2003proper}. Computing the full covariance matrix requires a Kronecker product between the spatial covariance matrix and the response covariance matrix. Consequently, the asymptotic computational time scales cubically with the number of locations and the number of responses. This approach is impractical in some applications where the geography is divided into thousands of regions, such as U.S. counties and tracts.

\citet{jin2005generalized} proposed a generalized multivariate CAR (GMCAR) model in which spatial random effects are defined through their marginal and conditional distributions. Each distribution contains a similar univariate spatial covariance matrix, that can effectively reduce both computational and storage demands.  Additionally, the GMCAR model improves estimation performance by allowing for different spatial autocorrelation parameters. Despite those advantages, the GMCAR model remains computationally inefficient for large spatial datasets. The running time for computing the inverse of the spatial precision matrix scales cubically with the number of locations. 

Dimension reduction through a truncated expansion of spatial basis functions is one common approach for reducing computational complexity in spatial modeling. For example, \citep{hughes2013dimension} employ Moran's I basis functions to estimate spatial generalized linear mixed models. \citep{janicki2022bayesian} utilize spatial basis functions to estimate a multivariate spatial mixture model. However, this dimension reduction approach has several criticisms. \citep{stein2014limitations} demonstrate the limitations of low-rank approximations using Kullback–Leibler divergence and numerical results. Additionally, \citep{bradley2016comparison} indicate that models with spatial basis functions tend to over-smooth the spatial process. This motivates our goal of developing more efficient techniques to model spatial structure without the need to take a reduced rank approach.


With recent development in machine learning, efficiently estimating model parameters has become feasible. Variational autoencoders (VAEs) are a class of neural network architectures that have emerged as a flexible and efficient approach for estimating probabilistic distributions and generating samples that resemble the observed data \citep{kingma2013auto, rezende2014stochastic, higgins2017beta, burgess2018understanding}. Building on these ideas, several efforts have been made to efficiently learn latent variable models using VAEs. \citet{zhang2023flexible} employ VAEs to estimate complex extreme-value processes in high-dimensional data. \citet{semenova2022priorvae} propose a two-stage procedure that bypasses the computational demands of latent Gaussian models by learning the spatial prior through a VAE. \citet{mishra2022pi} leverage VAEs to capture low-dimensional data representations. We note that neural networks have been used within a FH model (e.g., see \cite{parker2024nonlinear}), but not for the purpose of encoding spatial dependence.


In this paper, we develop models for multivariate small area estimation that effectively capture spatial dependencies. Specifically, we employ a $\beta$-VAE to effectively capture the complex neighborhood structures associated with area-level random effects, and seamlessly integrate this approach into the FH model. Once trained, the $\beta$-VAE serves as a generator with fixed parameters, adapting to any task involving any number of responses within the same geographic region. This approach greatly reduces computational overhead and improves scalability for large datasets. Although our interest in development of this methodology is for the purpose of small area estimation of official statistics, we note that the methodology itself is of independent interest to other domains that deal with spatially structured areal data, such as disease mapping, demography, and environmental statistics. The remainder of the paper is organized as follows. Section~\ref{sec:background} provides a brief overview of the background for the Multivariate Spatial FH model and VAE. In Section~\ref{sec:method}, we detail our proposed models for the FH model, addressing both separable and non-separable multivariate spatial random effects. Section~\ref{sec:sim} presents empirical simulations on both small-scale and large-scale datasets. In Section~\ref{sec:analysis}, we conduct a data analysis using all census tracts in California. Finally, we provide a brief summary in Section~\ref{sec:discussion}.

\section{Background}
\label{sec:background}
\subsection{Multivariate Spatial Fay-Herriot Model}
\label{sec:MS-FH}

In this section, we briefly review two commonly used multivariate spatial FH models. Detailed implementation are provided in \citep{porter2015small, jin2005generalized, fay1979estimates}. We begin by introducing the separable multivariate spatial FH model, denoted as SMS-FH. Consider the datasets with $N$ locations, $K$ responses and $P$ covariates. Let $\boldsymbol{Y}$ represent the $N \times K$ matrix of direct estimates, $\boldsymbol{\theta}$ be the $N \times K$ matrix representing the latent population quantities of interest, $\boldsymbol{X}$ be the $N \times P$ covariate matrix, and $\boldsymbol{\beta}$ be the $P \times K$ matrix of fixed effect coefficients. Additionally, $\boldsymbol{W}$ is the $N \times N$ adjacency matrix where diagonal entries are all zeros and non-zero off-diagonal entries indicate neighboring relationships, and $\boldsymbol{D}$ is the corresponding $N \times N$ diagonal degree matrix with diagonal elements $D_{ii} = \sum_j W_{ij}$. The SMS-FH model is defined as
\begin{eqnarray*}
    Y_{ik} &=& \theta_{ik} + \epsilon_{ik}, \qquad \epsilon_{ik} \sim N(0, \gamma^2_{ik}), \nonumber\\
    \theta_{ik} &=& \boldsymbol{X}_i \boldsymbol{\beta}_k + \phi_{ik} + \delta_{ik}, \qquad \delta_{ik} \sim N(0, \tau_k^2), \nonumber\\
    \text{vec}(\boldsymbol{\phi}) &\sim& N(\boldsymbol{0}, (\boldsymbol{D} - \rho\boldsymbol{W})^{-1} \otimes \boldsymbol{\Sigma}),\nonumber
\end{eqnarray*}
where \( \gamma_{ik} \) denotes the reported sampling error and \( \delta_{ik} \) captures the fine-scale variation for the \( k \)-th variable of the \( i \)-th area. \( \text{vec}(\cdot) \) is the vectorization operator. \( \boldsymbol{\phi} \) represents the $N \times K$ matrix of multivariate spatial random effect, which follows a separable MCAR distribution \citep{gelfand2003proper, jin2005generalized}. \( \boldsymbol{\Sigma} \) denotes the $K \times K$ correlation matrix between the multiple response variables. This model is referred to as separable because the correlation across variables is decoupled from the spatial covariance. Furthermore, each response marginally follows a CAR model and the spatial dependence parameter \( \rho \) is homogeneous across all responses, which limits the model's ability to capture complex multivariate relationships in area-level data.

\citet{jin2005generalized} introduced the GMCAR model that allows spatial autocorrelation to vary across different response variables, thus offering greater flexibility in modeling spatial dependencies. \citet{porter2015small} formally combined the GMCAR model with the FH model, resulting in what we refer to as the GMS-FH model. Here, we only introduce the GMS-FH model for bivariate responses. Extending it to cases with more than two response variables is relatively straightforward. Within the same framework, instead of sampling $\boldsymbol{\phi} = \{ \boldsymbol{\phi}_1, \boldsymbol{\phi}_2 \}$ jointly from the MCAR prior distribution, we sample $\boldsymbol{\phi}_2$ from its marginal distribution and then $\boldsymbol{\phi}_1$ from its conditional distribution given $\boldsymbol{\phi}_2$. The GMS-FH random effect structure is given as follows
\begin{eqnarray*}
    \boldsymbol{\phi}_1 | \boldsymbol{\phi}_2 &\sim& N(\boldsymbol{A}\boldsymbol{\phi}_2, \sigma^2_1 (\boldsymbol{D} - \rho_1\boldsymbol{W})^{-1}), \\
    \boldsymbol{\phi}_2 &\sim& N(\boldsymbol{0}, \sigma^2_2(\boldsymbol{D} - \rho_2\boldsymbol{W})^{-1}), \\
    \boldsymbol{A} &=& \eta_0 \boldsymbol{I} + \eta_1 \boldsymbol{W},
\end{eqnarray*}
where $\sigma_1$ and $\sigma_2$ are the scale parameters for the spatial random effects, while $\rho_1$ and $\rho_2$ represent the spatial dependence parameters. The terms $\eta_0$ and $\eta_1$ characterize the relationship between $\boldsymbol{\phi}_{1}$ and $\boldsymbol{\phi}_{2}$. Thus, the matrix $\boldsymbol{A}$ serves as a bridging matrix that connects $\boldsymbol{\phi}_1$ and $\boldsymbol{\phi}_2$. As a special case, SMS-FH can be represented by GMS-FH when both $\eta_0$ and $\eta_1$ are set to 0.


\subsection{Neural Networks}
In this section, we provide a brief introduction to neural networks. For more detailed information and advanced models, we refer the reader to \citet{Goodfellow-et-al-2016}. The simplest type of neural network is the  perceptron model \citep{rosenblatt1958perceptron}, which maps the input feature matrix $\boldsymbol{X}$ of size $N \times p$ to the output vector $f(\boldsymbol{X})$ of size $N$ using the following equation
\begin{align*}
  f(\boldsymbol{X}) = h(\boldsymbol{X} \cdot \boldsymbol{w}),
\end{align*}
where $\boldsymbol{w}$ represents the learned parameters of the perceptron, and $h(\cdot)$ is the activation function. In Rosenblatt's original work, a threshold function was used as the activation function to solve binary classification problems. Nowadays, modern neural networks typically use activation functions to introduce nonlinearity, thus increasing the model's capacity. In this context, any differentiable function can be used. For instance, the most widely used activation function is the Rectified Linear Unit (ReLU)
\begin{align*}
f(x) = \max(0, x),
\end{align*}
which is a piecewise linear function that returns 0 if the input is negative, otherwise returns the input itself. Although it is not differentiable at $x=0$, the standard practice in deep learning is to define the derivative as 0 at this point \citep{Goodfellow-et-al-2016}. It has become popular in deep learning due to its simplicity and efficient performance.

Feed-forward neural networks, also known as multi-layer perceptrons (MLPs), are the most fundamental deep network architecture. MLPs stack multiple layers of perceptrons by feeding the output of one perceptron as the input to the next, thereby further improving the model's ability to capture nonlinearities in the data. To find the optimal parameters, the gradient descent algorithm is typically used in MLPs. As all layers are differentiable, the back-propagation algorithm is employed to estimate the gradients efficiently. For a detailed discussion of optimization algorithms, we refer the reader to \citet{Goodfellow-et-al-2016}.

As a special neural network architecture, VAEs typically consist of  two symmetrical MLPs, denoted as the encoder and decoder networks. A detailed discussion of its theoretical foundations and its application in Bayesian inference can be found in Section \ref{sec:vae}.

\subsection{Variational Autoencoders}
\label{sec:vae}
VAEs typically serve two main purposes in statistical inferences. First, it provides a computationally efficient method for approximating posterior distributions in models with latent variables, which would otherwise be intractable or expensive to compute. Second, after learning the posterior distribution of the latent variables, the VAE can efficiently generate new samples that follow the distribution that resembles the input data \citep{kingma2013auto}.

Here, we employ the VAE as an efficient emulator for the prior distribution of spatial random effect $\boldsymbol{\phi}$. Specifically, we assume that the generation process of $\boldsymbol{\phi}$ is driven by an unknown latent random variable $\boldsymbol{z}$ with dimension $J$. Our objective is to jointly model the generative model $p_\zeta(\boldsymbol{\phi}, \boldsymbol{z})$ and an approximation to its posterior $p_\zeta(\boldsymbol{z} | \boldsymbol{\phi})$. The generative model can subsequently serve as an efficient substitute for generating the spatial prior in the context of Bayesian inference. 

Directly computing this posterior distribution is challenging because the marginal distribution \( p(\boldsymbol{\phi}) \) involves integrating over the entire latent space, which is often intractable or computationally expensive. VAE addresses this by introducing a probabilistic  encoder $q_\omega(\boldsymbol{z} | \boldsymbol{\phi})$ to approximate the posterior \( p_\zeta(\boldsymbol{z} | \boldsymbol{\phi}) \) and a probabilistic decoder $p_\zeta(\boldsymbol{\phi} | \boldsymbol{z})$ that reconstructs the input data $\boldsymbol{\phi}$ \citep{kingma2013auto}. The network parameters $\omega$ and $\zeta$ are typically given by two, often symmetric, neural networks. 

To optimize the parameters, \citet{kingma2013auto} employ the Stochastic Gradient Variational Bayes (SGVB) method to maximize the evidence lower bound (ELBO). \citet{higgins2017beta} proposed the $\beta$-VAE, which improves disentanglement in the latent space, and further enhance the performance of VAEs. The $\beta$-VAE incorporates an additional hyperparameter, denoted here as $\alpha$ to avoid confusion with coefficients associated with covariates.  This allows the ELBO to be defined as follows
\begin{equation}
    \mathcal{L}(\boldsymbol{\phi}) = - \alpha D_\text{KL}(q(\boldsymbol{z} | \boldsymbol{\phi}) \, || \, p(\boldsymbol{z})) + \mathbb{E}_{q(\boldsymbol{z} | \boldsymbol{\phi})} \left[ \log p(\boldsymbol{\phi} | \boldsymbol{z}) \right],
    \label{eq:vae_elbo}
\end{equation}
where the first term represents the negative KL divergence between the variational posterior of the latent variable and its prior. This term regularizes the learned latent variable $\boldsymbol{z}$ by aligning it with the prior distribution, ensuring a smooth and structured latent space while capturing information from the data. The second term corresponds to the expected log-likelihood, commonly referred to as the reconstruction error, which trains the decoder to map \(\boldsymbol{z}\) back to \(\tilde{\boldsymbol{\phi}}\), effectively reconstructing the input data. The parameter \(\alpha\) controls the trade-off between the reconstruction error and the KL divergence penalty. To simplify computation, \citet{kingma2013auto} assume a standard normal prior for \(\boldsymbol{z}\), which allows the KL divergence to be computed analytically. The reconstruction term is approximated using the Monte Carlo method. By incorporating these techniques, we obtain
\begin{equation*}
    \mathcal{L}(\boldsymbol{\phi}) \approx \frac{1}{2} \alpha \sum_{i=1}^M \sum_{j=1}^J \left(1 + \log\sigma_{ij}^2 - \mu_{ij}^2 - \sigma_{ij}^2\right) + \frac{1}{L} \sum_{i=1}^M \sum_{l=1}^L \log p(\boldsymbol{\phi}_{i} | \boldsymbol{z}^{(l)}_{i}),
\end{equation*}
where $M$ denotes the input batch size, $J$ is the latent dimension, and $L$ is the number of posterior samples for $\boldsymbol{z}$. The parameters $\sigma^2_{ij}$ and $\mu_{ij}$ of posterior distribution of $\boldsymbol{z}$ are obtained from the probabilistic encoder.


As demonstrated by \citet{semenova2022priorvae} and  \citet{mishra2022pi}, a VAE can be seamlessly integrated into Bayesian inference for area-level spatial models by replacing the computationally expensive spatial prior with the approximated prior generator learned by the VAE. Suppose the data of interest follows the likelihood \( p(\boldsymbol{y} \mid \boldsymbol{\xi}) \), parameterized by \(\boldsymbol{\xi}\). The goal of inference is to estimate the posterior distribution of \(\boldsymbol{\xi}\), which is given by Bayes' rule
\begin{align*}
    p(\boldsymbol{\xi} | \boldsymbol{y}) \propto p(\boldsymbol{y} | \boldsymbol{\xi})p(\boldsymbol{\xi}).
\end{align*}
The generative model learned by the VAE can serve as a replacement for the original prior. The resulting model is given by
\begin{eqnarray*}
    p(\tilde{\boldsymbol{\xi}} | \boldsymbol{y}) &\propto& p(\boldsymbol{y} | \tilde{\boldsymbol{\xi}})p(\tilde{\boldsymbol{\xi}}),\\
    \tilde{\boldsymbol{\xi}} &=& \text{decoder}(\boldsymbol{z}),
\end{eqnarray*}
where $\boldsymbol{z}$ is sampled from a standard normal distribution and $\text{decoder}(\cdot)$ is a MLP that serves as the generator for spatial random effects. This transformation converts the expensive computation into a simple sampling process from the standard normal prior followed by evaluating fixed deterministic functions of the VAE's decoder.



\section{Methodology}
\label{sec:method}
\subsection{Variational SMS-FH Model}
Consider again the SMS-FH model, the most computationally demanding step is the inverse of the spatial precision matrix. We aim to approximate the spatial random effect $\boldsymbol{\phi}$ with a VAE, thereby bypassing the need for this costly computation. As highlighted in \citet{semenova2022priorvae}, when modeling the random effect with a CAR prior, the VAE is more effective when it focuses solely on the spatial structure, excluding the covariance scale. To isolate the spatial component from the full covariance matrix, we express $\boldsymbol{\phi}$ using the matrix normal distribution
\begin{equation*}
    \boldsymbol{\phi} \sim \textit{MN}_{N \times K}(\boldsymbol{0}, (\boldsymbol{D} - \rho \boldsymbol{W}^{-1}), \boldsymbol{\Sigma}),
\end{equation*}
where $MN_{N \times K}$ denotes the matrix normal distribution with $N$ rows and $K$ columns, $\rho$ is the spatial autocorrelation parameter, and $\boldsymbol{\phi}$ is arranged with rows representing locations and columns representing variables. As in Section~\ref{sec:MS-FH}, \(\boldsymbol{D}\) and \(\boldsymbol{W}\) represent the degree matrix and adjacency matrix of the areal data, respectively, while \(\rho\) is the spatial dependence parameter. The matrix \(\boldsymbol{\Sigma}\) parameterizes the scale of the spatial covariance as well as the multivariate dependence among responses.

Let \( \boldsymbol{L} \) be the Cholesky decomposition of \( \boldsymbol{\Sigma} \), i.e., \( \boldsymbol{\Sigma} = \boldsymbol{L}\boldsymbol{L}^T \) and
\begin{equation}
    \boldsymbol{\psi} \sim \textit{MN}_{N \times K}(\boldsymbol{0}, (\boldsymbol{D} - \rho \boldsymbol{W})^{-1}, \boldsymbol{I}),
    \label{eq:VSMS-FH2}
\end{equation}
where $\boldsymbol{I}$ is the identity matrix of dimension $K \times K$. Using the properties of matrix normal random variable, we can write \( \boldsymbol{\phi} = \boldsymbol{\psi} \boldsymbol{L}^T \). This representation yields the factor $\boldsymbol{\psi}$, which captures only the spatial structure in the MCAR model. Following the VAE literature, we assume that the generating process of $\boldsymbol{\psi}$ depends on the variational latent variable $\boldsymbol{z}$. We can then train a VAE using samples drawn from (\ref{eq:VSMS-FH2}), and approximating $\boldsymbol{\psi}$ with $\tilde{\boldsymbol{\psi}} = \text{decoder}(\boldsymbol{z})$. This leads to the the variational SMS-FH model (VSMS-FH),
\begin{eqnarray}
    Y_{ik} &=& \theta_{ik} + \epsilon_{ik},\nonumber\\
    \theta_{ik} &=& \boldsymbol{X}_i \boldsymbol{\beta}_k + \tilde{\phi}_{ik} + \delta_{ik}, \nonumber
\end{eqnarray}
where $\tilde{\boldsymbol{\phi}}$ is represented as
\begin{eqnarray}
    \tilde{\boldsymbol{\phi}} &=& \sigma \tilde{\boldsymbol{\psi}}, \nonumber\\
    \tilde{\boldsymbol{\psi}} &=& \text{decoder}(\boldsymbol{z}),\nonumber
\end{eqnarray}
where $\boldsymbol{z}$ follows a standard normal distribution. Lastly, we note that any number of samples can be drawn from the distribution of $\boldsymbol{\psi}$ to train the VAE. The quality of the random effects generated by the VAE decoder improves as the size of the training dataset increases \citep{semenova2022priorvae}.



  \subsection{Variational GMS-FH Model}
As noted in Section~\ref{sec:MS-FH}, the SMS-FH model is less flexible due to the separability across variables and the homogeneity in spatial autocorrelation. \citet{porter2015small} argues that such assumptions are plausible primarily when survey responses are demographically related. In other scenarios, these assumptions may not hold. In this section, we introduce the VGMS-FH model to address this issue.

We again consider the case with two response variables, with the extension to more than two being straightforward. It is important to note that the distributions for \( \boldsymbol{\phi}_2 \) and \( \boldsymbol{\phi}_1 | \boldsymbol{\phi}_2 \) exhibit the similar underlying covariance structure. Accordingly, they can be represented concisely as the following
\begin{eqnarray}
    \boldsymbol{\phi}_2 &=& \sigma_2 \boldsymbol{\psi}_2, \nonumber\\
    \boldsymbol{\phi}_1 &=& \boldsymbol{A} \boldsymbol{\phi}_2 + \sigma_1 \boldsymbol{\psi}_1, \nonumber\\
    \boldsymbol{\psi}_i &\sim& N(\boldsymbol{0}, (\boldsymbol{D} - \rho_i \boldsymbol{W})^{-1}), \label{eq:VGMS-FH3}
\end{eqnarray}
where $\{\rho_i: i = 1, 2\}$ denotes the spatial dependence parameters, each with the same prior distribution. Recall that all information from the spatial prior, including the prior for the spatial dependence parameter, can be approximated by a VAE. Since $\rho_i$ shares the same prior, we can reuse the same VAE trained with a univariate CAR prior to model both the marginal and conditional distributions of $\boldsymbol{\phi}$. Specifically, we can draw as much training data as needed from (\ref{eq:VGMS-FH3}) and fit a VAE to learn $\tilde{\boldsymbol{\psi}}_i = \text{decoder}(\boldsymbol{z})$. The VGMS-FH model can then be formulated as
\begin{eqnarray*}
    Y_{ik} &=& \theta_{ik} + \epsilon_{ik},\\
    \theta_{ik} &=& \boldsymbol{X}_i \boldsymbol{\beta}_k + \tilde{\phi}_{ik} + \delta_{ik},
\end{eqnarray*}
where, for $\boldsymbol{z}_1, \boldsymbol{z}_2 \sim N(\boldsymbol{0}, \boldsymbol{1})$, $\tilde{\boldsymbol{\phi}}$ can be approximated as
\begin{eqnarray*}
    \tilde{\boldsymbol{\phi}}_2 &=& \sigma_2 \tilde{\boldsymbol{\psi}}_2,\\
    \tilde{\boldsymbol{\phi}}_1 &=& \boldsymbol{A} \tilde{\boldsymbol{\phi}}_2 + \sigma_1 \tilde{\boldsymbol{\psi}}_1,\\
    \tilde{\boldsymbol{\psi}}_2 &=& \text{decoder}(\boldsymbol{z}_2),\\
    \tilde{\boldsymbol{\psi}}_1 &=& \text{decoder}(\boldsymbol{z}_1).
\end{eqnarray*}
It is worth emphasizing that the function \( \text{decoder}(\cdot) \) involves only matrix multiplication and the element-wise applying of nonlinear functions, both of which are computationally inexpensive. For this reason, the proposed VGMS-FH model is significantly more efficient than the GMS-FH model and can scale to datasets with much higher-dimensional geographies. In addition, VGMS-FH tends to be slightly faster than VSMS-FH, despite being more flexible. This increased efficiency in VGMS-FH results from avoiding the need to sample the covariance matrix of responses, typically from the Inverse Wishart distribution, and perform subsequent Cholesky decomposition. The simulation study in Section~\ref{sec:sim} also confirms this conclusion.

Finally, it should be noted that in both VSMS-FH and VGMS-FH, the process of drawing training data from the CAR prior still involves the computationally expensive matrix inverse, and training a VAE to generate spatial random effects can also be time-consuming. However, these steps only need to be performed once, given that the geographical boundaries remain unchanged for a long period (e.g., census tract definitions are often revised during the decennial census). Consequently, the trained VAE can be considered fixed and subsequently applied to any analysis requiring a spatial random effect, without the need for retraining.

\section{Empirical Simulation Study}
\label{sec:sim}

To evaluate the performance of the proposed model, we conduct two empirical simulations, focusing on scenarios with both low-dimensional and high-dimensional geographies. For the low-dimensional simulation, we use data on median household income, median monthly housing costs, and their corresponding margins of error for all 115 counties in Missouri. For the high-dimensional simulation, we use the same variables for 8893 census tracts in California, excluding tracts that were removed due to isolated regions and missing observations in both responses and covariates. The data for both simulations are sourced from the 5-year American Community Survey (ACS) released in 2020 \citep{acs_data, tidycensus}. We apply a log transformation to better approximate the normality assumption and use the Delta method to calculate the corresponding variance. For the purposes of simulation, the log-transformed values of the two variables are treated as the true population values, and noisy samples (simulated ``direct estimates") are generated by adding noise drawn from the reported sampling errors around these population values. Additionally, we include the poverty rate, and four demographic variables (proportions of Black, Asian, and American Indian and Alaska Native populations) as covariates.

\subsection{Model Implementation}
\label{sec:implementation}
We follow the same two-step procedure for estimating both the VSMS-FH and VGMS-FH models as described in Section~\ref{sec:method}. In the first step, we employ a VAE to learn the spatial random effect $\boldsymbol{\phi}$ using 10,000 samples drawn from the spatial prior. For the SMS-FH model, $\boldsymbol{\phi}$ is sampled from $N(\boldsymbol{0}, (\boldsymbol{D} - \rho \boldsymbol{W})^{-1} \otimes \boldsymbol{I})$, where $\rho \sim \text{Unif}(0, 1)$. Similarly, for the GMS-FH model, $\boldsymbol{\phi}$ is sampled from $N(\boldsymbol{0}, (\boldsymbol{D} - \rho \boldsymbol{W})^{-1})$, with $\rho \sim \text{Unif}(0, 1)$. Both the encoder and decoder of the VAE are implemented with one hidden layer and one latent layer. The activation function in these neural netwoks is the exponential linear unit (ELU) function. Following \citet{semenova2022priorvae}, we set the dimensions of the hidden and latent layers equal to those of the input layer, as the sampled prior contains only moderate information. Additionally, we adopt a $\beta$-VAE structure to improve input reconstruction, with $\beta$ set to the inverse of the latent dimension. 

In the second step, we apply Hamiltonian Monte Carlo (HMC) to estimate the FH model with spatial random effects generated by the trained decoders. We assign uninformative priors to reflect our limited knowledge of the parameters. For the SMS-FH model, we specify $\tau^2_k \sim IG(0.001, 0.001)$ and $\beta_{pk} \sim N(0, 100)$. For the GMS-FH model, we set priors for both $\gamma_1^2$ and $\gamma_2^2$ as $IG(0.001, 0.001)$, and for both $\eta_0$ and $\eta_1$ as $N(0, 100)$, following the recommendations in \citet{jin2005generalized} and \citet{porter2015small}.

For comparison, we evaluate the direct estimates, as well as the multivariate FH, SMS-FH, GMS-FH, VSMS-FH, and VGMS-FH models. The VAE for generating spatial random effects is implemented using PyTorch \citep{paszke2017automatic} and ran on the high-performance computing GPU cluster at the University of Missouri, which has four Nvidia A100 GPUs, each with 80GB of RAM. We utilized one of these GPUs and allocated 16GB for training VAE. The subsequent HMC steps are carried out with the GPU-accelerated probabilistic programming library, NumPyro \citep{phan2019composable}. We note that the HMC step can also be implemented on using any software preferred by the user, such as Stan \citep{carpenter2017stan}. We run the HMC on both CPU and GPU clusters for 20,000 simulations with first 10,000 simulations as burn-in. Our CPU cluster contains 2 AMD 7713 processors and 512GB of RAM. We allocated 64GB for training HMC. For the GPU cluster, we allocated 30GB for this task. We evaluate model performance using three metrics: root mean square error (RMSE), interval score (IS) \citep{gneiting2007strictly}, and mean coverage rate (Coverage).

\subsection{Low-Dimensional Simulation}
For the low-dimensional simulation, we repeat the above procedure 100 times to create 100 simulated datasets. In each dataset, we fit all five models and compare their performance in Table~\ref{tab:sim_mo}.
\begin{table}[H]
\centering
\caption{Performance metrics on the original scale for 5-year period county-level estimates in Missouri obtained from the 2020 American Community Survey. All metrics are calculated against the true population values.}
\label{tab:sim_mo}
\begin{tabular}[H]{lcccc}
\toprule
Model                  & RMSE      &   IS     & Coverage & CPU Time\\
\midrule
\multicolumn{5}{c}{Median Household Income} \\
\midrule
Direct Estimate        & 2314.875 & 11001.929& 0.950 & NA\\
FH                     & 1942.552 & 9773.144 & 0.941 & 1.08 mins\\
SMS-FH                 & 1861.450 & 8936.116 & 0.953 & 12.72 hours \\
VSMS-FH                & 1866.506 & 8967.448 & 0.952 & 7.88 mins\\
GMS-FH                 & 1803.185 & 9251.591 & 0.932 & 3.14 hours\\
VGMS-FH                & 1808.740 & 9138.526 & 0.934 & 6.47 mins\\
\midrule
\multicolumn{5}{c}{Median Monthly Housing Costs} \\
\midrule
Direct Estimate        & 22.613   & 107.000  & 0.952 & NA\\
FH                     & 21.540   & 101.308  & 0.952 & 1.08 mins\\
SMS-FH                 & 21.014   & 98.875   & 0.953 & 12.72 hours\\
VSMS-FH                & 21.033   & 98.907   & 0.953 & 7.88 mins\\
GMS-FH                 & 20.751   & 97.487   & 0.951 & 3.14 hours\\
VGMS-FH                & 20.670   & 96.887   & 0.953 & 6.47 mins\\
\bottomrule
\end{tabular}
\end{table}

We observe that the FH model without spatial random effects performs worse across all metrics compared to models with spatial random effects, confirming that incorporating spatial random effects is beneficial for small area estimation. Furthermore, the GMCAR based models clearly outperform the separable MCAR based models in terms of RMSE, though they appear to perform less well on interval estimations. This underperformance is primarily due to poor results in Hickory County. Excluding this county, the interval scores are as follows: SMS-FH at 8749.671, VSMS-FH at 8754.984, GMS-FH at 8634.957, and VGMS-FH at 8577.589. This confirms that the GMCAR more effectively captures the true population values, as discussed in Section~\ref{sec:method}. Finally, we note that both the VMS-FH and VGMS-FH models show comparable performance to their full HMC counterparts, SMS-FH and GMS-FH, respectively. This highlights the effectiveness of the VAE in modeling spatial random effects. The main advantage of the VAE-based models lies in their computational efficiency. On the CPU cluster, the GMS-FH model requires 3.14 hours per simulation, while the VGMS-FH model completes the same task in just 6.47 minutes.

\subsection{High-dimensional Simulation}
\label{sec:sim_large}
For the high-dimensional simulation, we only compare the VAE-based model to the direct estimate and the FH model without spatial random effects on 20 simulated datasets. Both the SMS-FH and GMS-FH models are too computationally burdensome to be fit at this scale. For example, it is estimated that GMS-FH requires over 45 days to complete a single simulation, making it impractical for this task. It is worth noting that area-level spatial models without any form of dimension reduction are generally not applied to such high-dimensional geographical areas. However, we report the GMS-FH's estimated running time to highlight the efficiency of the proposed model. The performance metrics of VSMS-FH and VGMS-FH are presented in Table~\ref{tab:sim_ca}. Coverage rate metrics are omitted in this scenario due to the limited number of simulated datasets.
\begin{table}[H]
\centering
\caption{Performance metric on the original scale for 5-year period tract-level estimates in California obtained from the 2020 American Community Survey. All metrics are calculated against the true population values.}
\label{tab:sim_ca}
\begin{tabular}[H]{lcc}
\toprule
Model                        & RMSE        &   IS \\
\midrule
\multicolumn{3}{c}{Median Household Income} \\
\midrule
Direct Estimate        & 11281.511 & 57241.784\\
FH                     & 9449.140  & 46891.550\\
VSMS-FH                &  8944.752  & 44635.729\\
VGMS-FH                & 8674.416  & 43816.726\\
\midrule
\multicolumn{3}{c}{Median Monthly Housing Costs} \\
\midrule
Direct Estimate        & 138.365   &  680.593 \\
FH                     & 127.816   &  630.843\\
VSMS-FH                &  125.324  & 624.411\\
VGMS-FH                & 124.483   &  622.895\\
\bottomrule
\end{tabular}
\end{table}

The results show a clear performance improvement of the spatial models over both the direct estimates and the FH model. Consistent with our previous finding, VGMS-FH slightly outperforms VSMS-FH in terms of both RMSE and IS. Most notably, the VAE-based model makes fitting a multivariate spatial FH model feasible on this high-dimensional geographical dataset. Each simulation of VGMS-FH takes approximately 2.80 hours on the GPU cluster, and VSMS-FH takes about 3.51 hours.

\section{American Community Survey Analysis}
\label{sec:analysis}
Modeling household income and housing costs is essential for informed policy-making and regional planning, as these two factors together offer critical insights into economic well-being and affordability within communities. For instance, \citet{wardrip2008fully} evaluates the burden of housing costs on renters and homeowners. \citet{joice2014measuring} uses household income data from the ACS to assess housing affordability. Here, we apply the proposed model to jointly estimate the population values for household income and housing costs. Direct estimates and the same covariates described in Section~\ref{sec:sim} were obtained from the ACS data for 9,040 census tracts in California. While California contains 9,129 tracts, 20 were excluded due to missing geographic information, 6 were omitted as they were identified as isolated, and 63 were excluded due to missing covariate data. After removing those records, 147 tracts still have missing values in the direct estimate or margin of errors for the response variables. Rather than removing these missing responses as in Section~\ref{sec:sim}, we use spatial interpolation to impute them. Our exploratory analysis indicates a positive correlation between housing costs and household income. A thorough description of the data is available in Section~\ref{sec:sim}. 

\subsection{Estimation Results}
We follow the same implementation details for the VGMS-FH model as outlined in Section~\ref{sec:implementation}. To handle missing observations in the estimated population parameters, we interpolate them as follows  
\begin{equation*}
    \text{vec}(\boldsymbol{Y}) = \boldsymbol{H} \, \text{vec}(\boldsymbol{\theta}) + \text{vec}(\boldsymbol{\epsilon}),
    \label{eq:interpolation}
\end{equation*}  
where $\boldsymbol{H}$ is the incidence matrix of size $M \times NP$, with $M$ representing the length of $\text{vec}(\boldsymbol{Y})$. For further details on spatial interpolation, we refer readers to \citet{wikle2019spatio}.

\begin{figure}[H]
\begin{center}
\includegraphics[width=0.75\textwidth,keepaspectratio]{"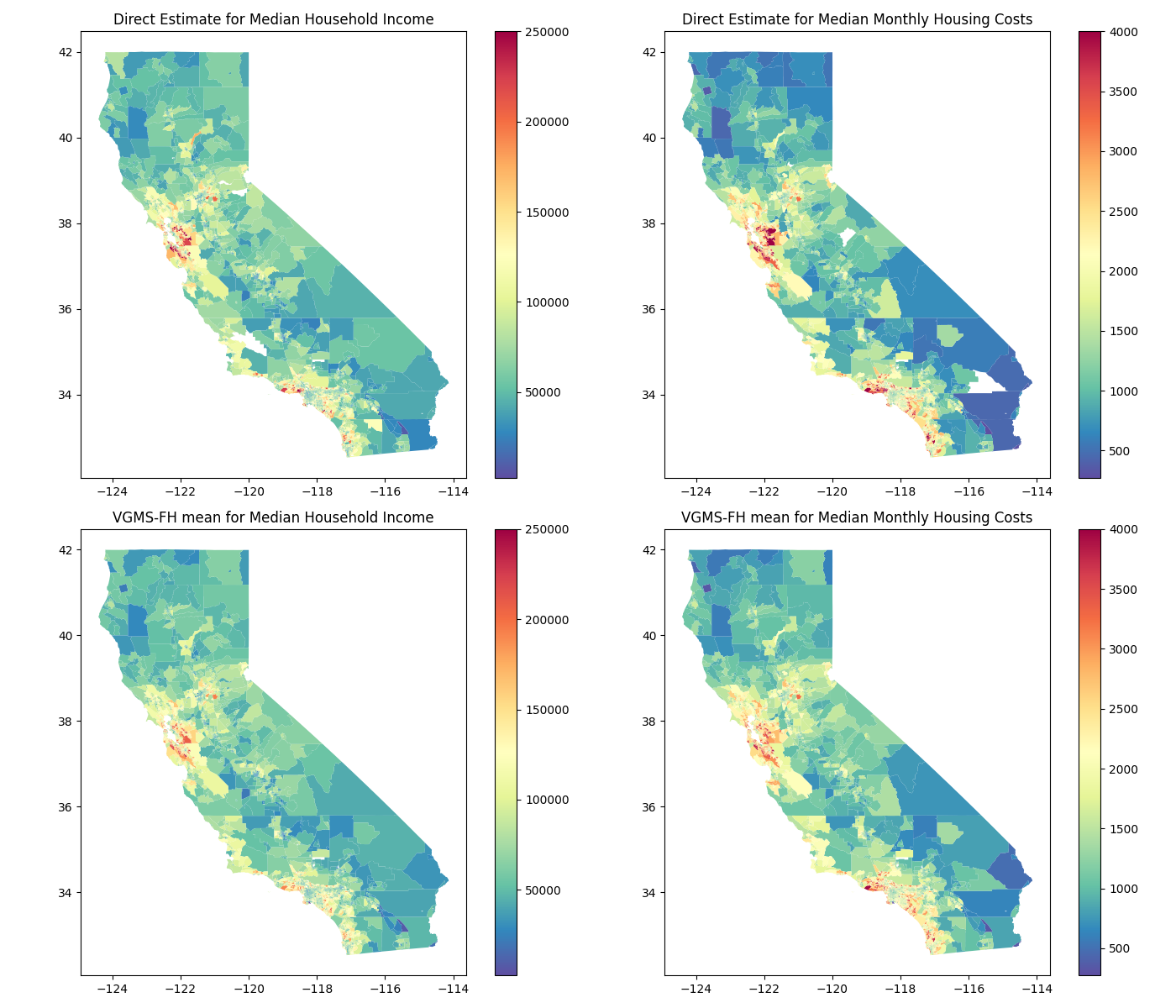"}
\caption{Estimates of median household income and median monthly housing costs for 9,040 census tracts in California, based on 5-year period estimates from the 2020 American Community Survey, including both direct estimates and results from the VGMS-FH model}
\label{fig:vae_mean}
\end{center}
\end{figure}

In Figure~\ref{fig:vae_mean}, we present the posterior mean values from the VGMS-FH model along with the direct estimates. The VGMS-FH results reveal consistent spatial patterns, with both household income and housing costs generally higher on the West Coast and lower in the northern and southeastern regions. Additionally, the VGMS-FH model effectively handles missing values for both responses by generating interpolated estimates that incorporate both covariates and the underlying spatial structure.

\begin{figure}[H]
\begin{center}
\includegraphics[width=0.75\textwidth,keepaspectratio]{"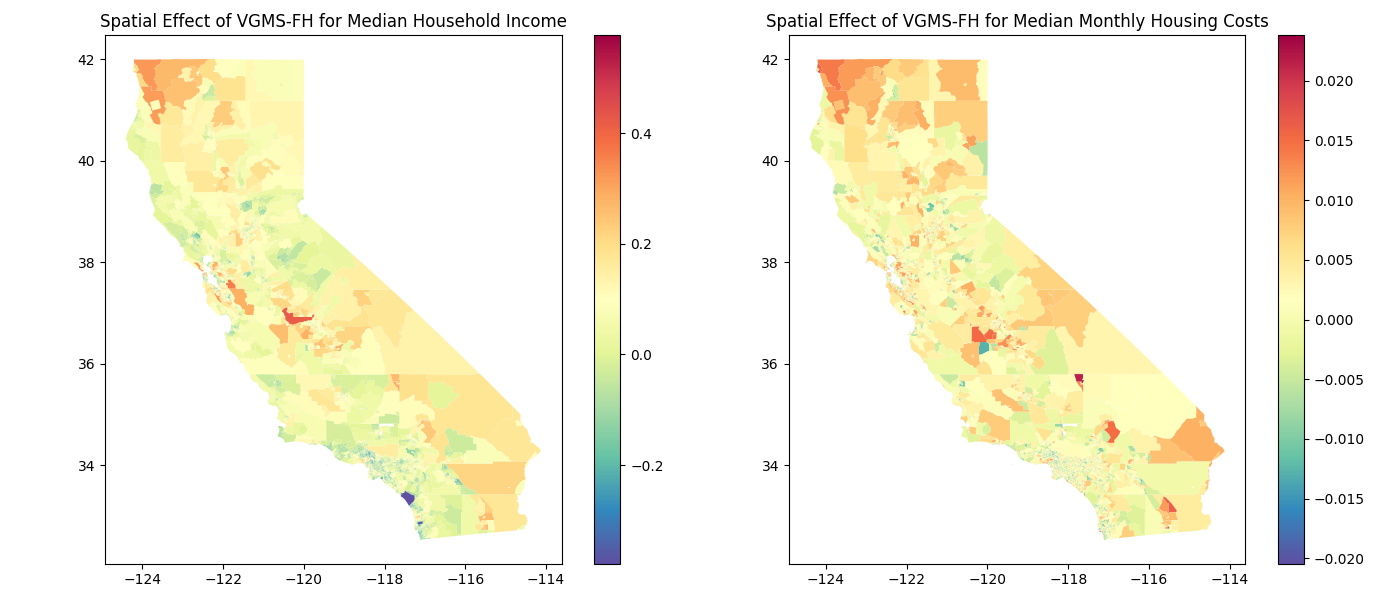"}
\caption{Spatial random effects on the log-scale of median household income and median monthly housing costs for 9,040 census tracts in California, based on 5-year period estimates from the 2020 American Community Survey, including both direct estimates and results from the VGMS-FH model}
\label{fig:vae_car}
\end{center}
\end{figure}

Figure~\ref{fig:vae_car} displays the spatial random effects learned by the VAE. The results indicate that the covariates capture much of the overall spatial variation, while the spatial random effects account for remaining spatial structure, particularly in the northern regions. By modeling these finer spatial dependencies, the VGMS-FH model improves the accuracy of the estimates, especially in areas where the covariates alone do not fully explain the observed variation.

\begin{figure}[H]
\begin{center}
\includegraphics[width=0.75\textwidth,keepaspectratio]{"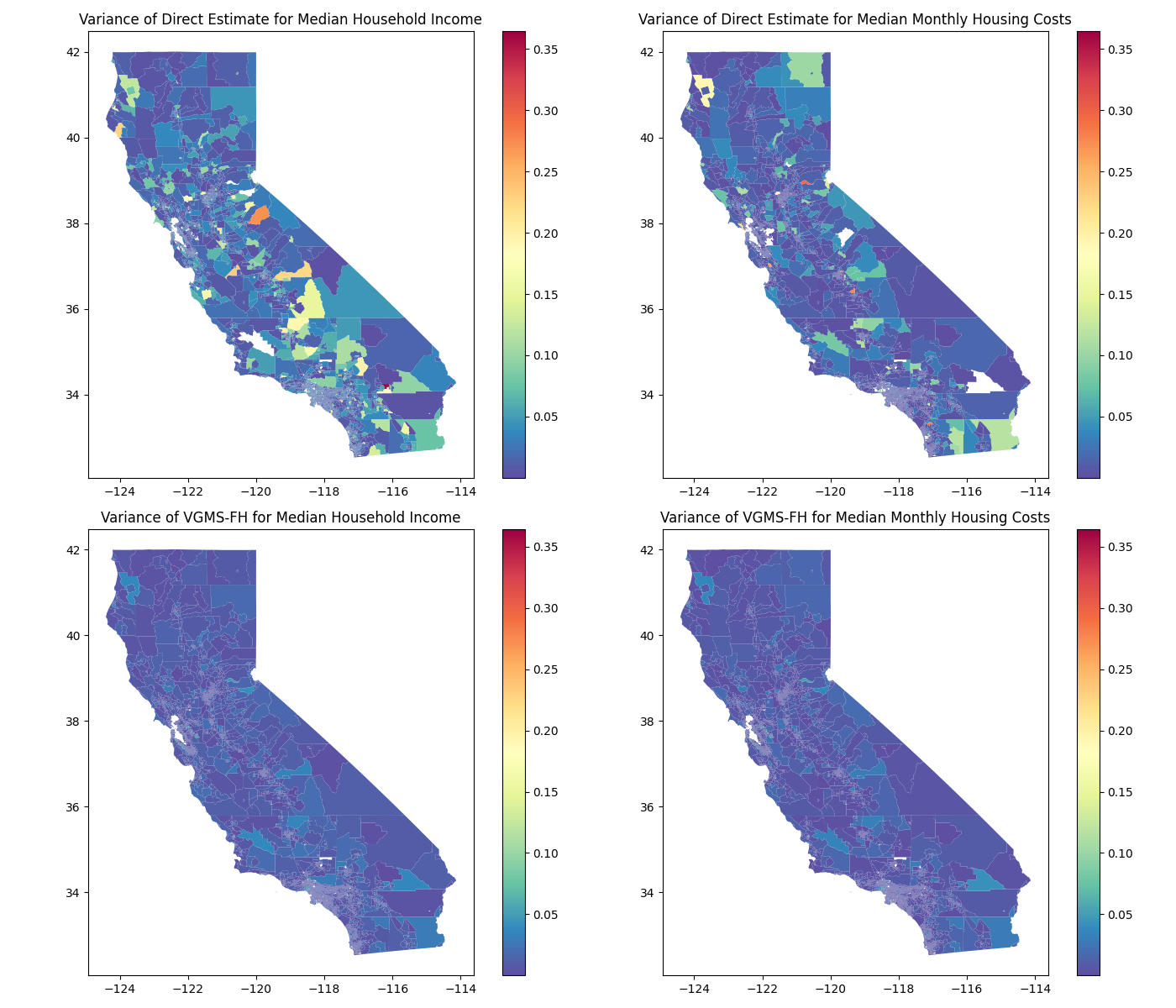"}
\caption{Variance on the log-scale of median household income and median monthly housing costs for 9,040 census tracts in California, based on 5-year period estimates from the 2020 American Community Survey, including both direct estimates and results from the VGMS-FH model.}
\label{fig:vae_var}
\end{center}
\end{figure}

A comparison of log-scale variance between direct estimates and the VGMS-FH model is provided in Figure~\ref{fig:vae_var}. The VGMS-FH model achieves a substantial reduction in variance across most regions, increasing the stability and reliability of the estimates. This improvement is particularly notable in areas with limited sample sizes or high variability.

\begin{figure}[H]
\begin{center}
\includegraphics[width=0.75\textwidth,keepaspectratio]{"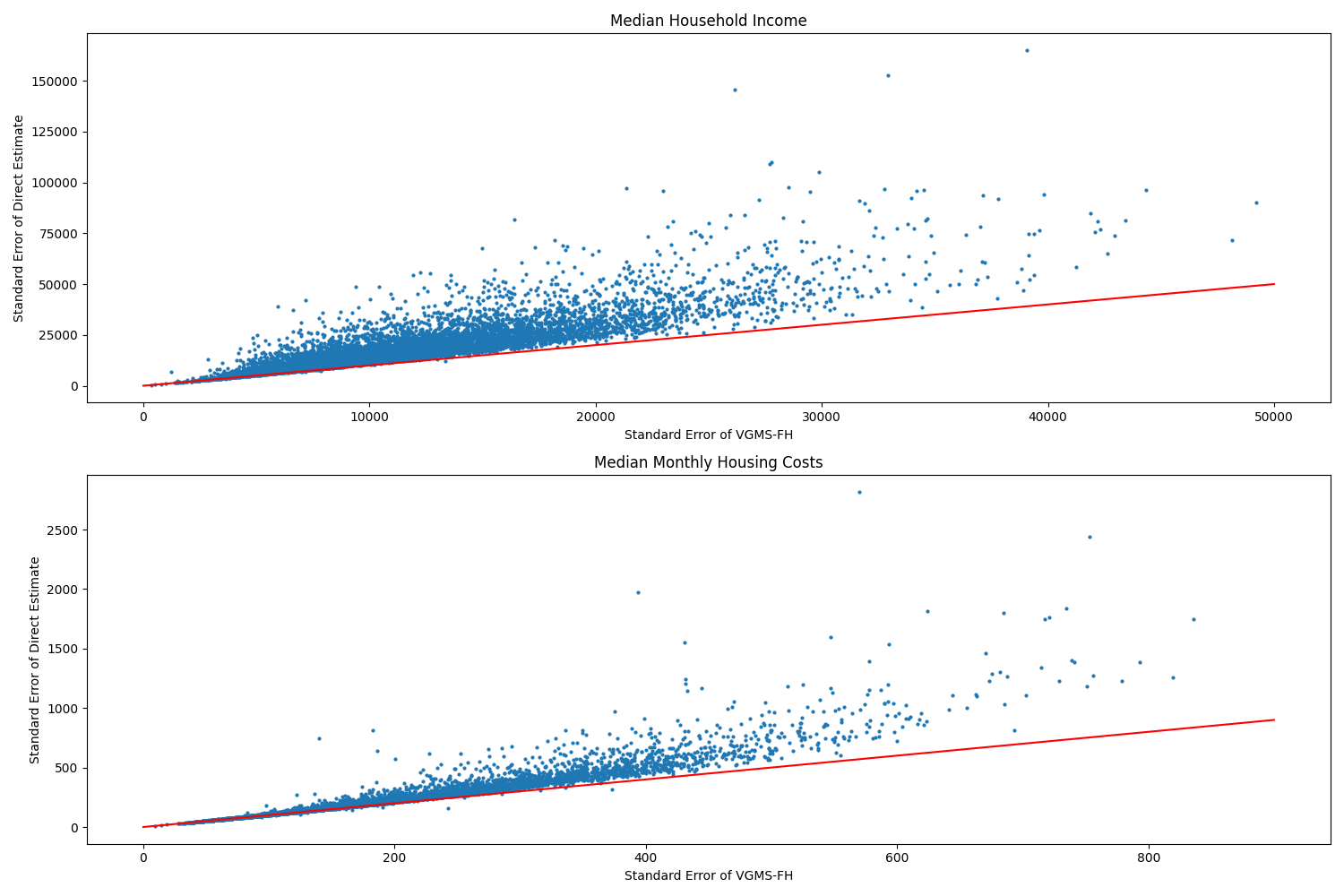"}
\caption{Standard error on the original scale of direct estimate versus results of VGMS-FH model, based on 5-year period estimates from the 2020 American Community Survey data.}
\label{fig:vae_var_scatter}
\end{center}
\end{figure}

Finally, Figure~\ref{fig:vae_var_scatter} displays scatter plots comparing the standard errors of direct estimates and VGMS-FH model estimates on the original scale. The VGMS-FH model significantly reduces the standard error across nearly all regions, indicating a substantial improvement over direct estimates. Furthermore, the scatter plots indicate that regions with higher sampling errors exhibit more substantial improvements with the VGMS-FH model.

\section{Discussion}
\label{sec:discussion}
In small area estimation, incorporating spatial structure allows regions with fewer samples to borrow strength from neighboring regions. However, integrating spatial dependence into the FH model requires substantial computational resources, which limits the model’s capacity for high-dimensional geographic areas. This issue becomes particularly pronounced when handling multiple responses, as the computational cost for sampling from the MCAR model increases cubically.

In this paper, we use the VAE to approximate the prior distribution of the multivariate spatial random effects and embed this within the FH model. We demonstrate that the VAE effectively captures complex neighborhood structures in area-level data and provides accurate estimates for small area estimation. Among the two models, we show that the VGMS-FH model can capture neighborhood dependencies across different responses using a unified covariance structure. This framework not only reduces computational costs but also allows the spatial structure to be applied to any analysis within the same geographic regions. Through empirical simulation studies, we shows that the VAE can scale effectively to high-dimensional geographies, such as census tracts in California. The efficiency and adaptability of our proposed method make it particularly well-suited for high-dimensional datasets in official statistics. 

\section*{Acknowledgments}
  This article is released to inform interested parties of ongoing research and to encourage discussion. The views expressed on statistical issues are those of the authors and not those of the NSF or U.S. Census Bureau. This research was partially supported by the U.S. National Science Foundation (NSF) under NSF grants NCSE-2215168 and NCSE-2215169. The authors thank Jerry Maples and Sallie Keller for helpful comments on an earlier version of this paper.

\bibliography{reference}

\begin{thebibliography}{30}
\newcommand{\enquote}[1]{``#1''}
\expandafter\ifx\csname natexlab\endcsname\relax\def\natexlab#1{#1}\fi

\bibitem[\protect\citename{Bradley et~al., }2016]{bradley2016comparison}
Bradley, J.~R., Cressie, N., and Shi, T. (2016).
\newblock \enquote{A comparison of spatial predictors when datasets could be very large.}

\bibitem[\protect\citename{Burgess et~al., }2018]{burgess2018understanding}
Burgess, C.~P., Higgins, I., Pal, A., Matthey, L., Watters, N., Desjardins, G., and Lerchner, A. (2018).
\newblock \enquote{Understanding disentangling in $\beta$-VAE.}
\newblock {\em arXiv preprint arXiv:1804.03599\/}.

\bibitem[\protect\citename{Carpenter et~al., }2017]{carpenter2017stan}
Carpenter, B., Gelman, A., Hoffman, M.~D., Lee, D., Goodrich, B., Betancourt, M., Brubaker, M., Guo, J., Li, P., and Riddell, A. (2017).
\newblock \enquote{Stan: A probabilistic programming language.}
\newblock {\em Journal of statistical software\/}, 76, 1--32.

\bibitem[\protect\citename{Chandra et~al., }2015]{chandra2015spatially}
Chandra, H., Salvati, N., and Chambers, R. (2015).
\newblock \enquote{A spatially nonstationary Fay-Herriot model for small area estimation.}
\newblock {\em Journal of Survey Statistics and Methodology\/}, 3, 2, 109--135.

\bibitem[\protect\citename{Fay~III and Herriot, }1979]{fay1979estimates}
Fay~III, R.~E. and Herriot, R.~A. (1979).
\newblock \enquote{Estimates of income for small places: an application of James-Stein procedures to census data.}
\newblock {\em Journal of the American Statistical Association\/}, 74, 366a, 269--277.

\bibitem[\protect\citename{Gelfand and Vounatsou, }2003]{gelfand2003proper}
Gelfand, A.~E. and Vounatsou, P. (2003).
\newblock \enquote{Proper multivariate conditional autoregressive models for spatial data analysis.}
\newblock {\em Biostatistics\/}, 4, 1, 11--15.

\bibitem[\protect\citename{Gneiting and Raftery, }2007]{gneiting2007strictly}
Gneiting, T. and Raftery, A.~E. (2007).
\newblock \enquote{Strictly proper scoring rules, prediction, and estimation.}
\newblock {\em Journal of the American statistical Association\/}, 102, 477, 359--378.

\bibitem[\protect\citename{Goodfellow et~al., }2016]{Goodfellow-et-al-2016}
Goodfellow, I., Bengio, Y., and Courville, A. (2016).
\newblock {\em Deep Learning\/}.
\newblock MIT Press.
\newblock \url{http://www.deeplearningbook.org}.

\bibitem[\protect\citename{Higgins et~al., }2017]{higgins2017beta}
Higgins, I., Matthey, L., Pal, A., Burgess, C.~P., Glorot, X., Botvinick, M.~M., Mohamed, S., and Lerchner, A. (2017).
\newblock \enquote{beta-vae: Learning basic visual concepts with a constrained variational framework.}
\newblock {\em ICLR (Poster)\/}, 3.

\bibitem[\protect\citename{Hughes and Haran, }2013]{hughes2013dimension}
Hughes, J. and Haran, M. (2013).
\newblock \enquote{Dimension reduction and alleviation of confounding for spatial generalized linear mixed models.}
\newblock {\em Journal of the Royal Statistical Society Series B: Statistical Methodology\/}, 75, 1, 139--159.

\bibitem[\protect\citename{Janicki et~al., }2022]{janicki2022bayesian}
Janicki, R., Raim, A.~M., Holan, S.~H., and Maples, J.~J. (2022).
\newblock \enquote{Bayesian nonparametric multivariate spatial mixture mixed effects models with application to American Community Survey special tabulations.}
\newblock {\em The Annals of Applied Statistics\/}, 16, 1, 144--168.

\bibitem[\protect\citename{Jin et~al., }2005]{jin2005generalized}
Jin, X., Carlin, B.~P., and Banerjee, S. (2005).
\newblock \enquote{Generalized hierarchical multivariate CAR models for areal data.}
\newblock {\em Biometrics\/}, 61, 4, 950--961.

\bibitem[\protect\citename{Joice, }2014]{joice2014measuring}
Joice, P. (2014).
\newblock \enquote{Measuring housing affordability.}
\newblock {\em Cityscape\/}, 16, 1, 299--308.

\bibitem[\protect\citename{Kingma, }2013]{kingma2013auto}
Kingma, D. (2013).
\newblock \enquote{Auto-Encoding Variational Bayes.}
\newblock {\em arXiv preprint arXiv:1312.6114\/}.

\bibitem[\protect\citename{Mishra et~al., }2022]{mishra2022pi}
Mishra, S., Flaxman, S., Berah, T., Zhu, H., Pakkanen, M., and Bhatt, S. (2022).
\newblock \enquote{$\pi$ VAE: a stochastic process prior for Bayesian deep learning with MCMC.}
\newblock {\em Statistics and Computing\/}, 32, 6, 96.

\bibitem[\protect\citename{Parker, }2024]{parker2024nonlinear}
Parker, P.~A. (2024).
\newblock \enquote{Nonlinear Fay-Herriot Models for Small Area Estimation Using Random Weight Neural Networks.}
\newblock {\em Journal of Official Statistics\/},  0282423X241244671.

\bibitem[\protect\citename{Paszke et~al., }2017]{paszke2017automatic}
Paszke, A., Gross, S., Chintala, S., Chanan, G., Yang, E., DeVito, Z., Lin, Z., Desmaison, A., Antiga, L., and Lerer, A. (2017).
\newblock \enquote{Automatic differentiation in PyTorch.}

\bibitem[\protect\citename{Phan et~al., }2019]{phan2019composable}
Phan, D., Pradhan, N., and Jankowiak, M. (2019).
\newblock \enquote{Composable Effects for Flexible and Accelerated Probabilistic Programming in NumPyro.}
\newblock {\em arXiv preprint arXiv:1912.11554\/}.

\bibitem[\protect\citename{Porter et~al., }2014]{porter2014spatial}
Porter, A.~T., Holan, S.~H., Wikle, C.~K., and Cressie, N. (2014).
\newblock \enquote{Spatial Fay--Herriot models for small area estimation with functional covariates.}
\newblock {\em Spatial Statistics\/}, 10, 27--42.

\bibitem[\protect\citename{Porter et~al., }2015]{porter2015small}
Porter, A.~T., Wikle, C.~K., and Holan, S.~H. (2015).
\newblock \enquote{Small area estimation via multivariate Fay--Herriot models with latent spatial dependence.}
\newblock {\em Australian \& New Zealand Journal of Statistics\/}, 57, 1, 15--29.

\bibitem[\protect\citename{Rezende et~al., }2014]{rezende2014stochastic}
Rezende, D.~J., Mohamed, S., and Wierstra, D. (2014).
\newblock \enquote{Stochastic backpropagation and approximate inference in deep generative models.}
\newblock In {\em International conference on machine learning\/},  1278--1286. PMLR.

\bibitem[\protect\citename{Rosenblatt, }1958]{rosenblatt1958perceptron}
Rosenblatt, F. (1958).
\newblock \enquote{The perceptron: a probabilistic model for information storage and organization in the brain.}
\newblock {\em Psychological review\/}, 65, 6, 386.

\bibitem[\protect\citename{Salvati, }2004]{salvati2004small}
Salvati, N. (2004).
\newblock \enquote{Small area estimation by spatial models: the spatial empirical best linear unbiased prediction (spatial EBLUP).}
\newblock {\em Dipartimento di Statistica” G. Parenti” viale morgagni\/},  59--50134.

\bibitem[\protect\citename{Semenova et~al., }2022]{semenova2022priorvae}
Semenova, E., Xu, Y., Howes, A., Rashid, T., Bhatt, S., Mishra, S., and Flaxman, S. (2022).
\newblock \enquote{PriorVAE: encoding spatial priors with variational autoencoders for small-area estimation.}
\newblock {\em Journal of the Royal Society Interface\/}, 19, 191, 20220094.

\bibitem[\protect\citename{Stein, }2014]{stein2014limitations}
Stein, M.~L. (2014).
\newblock \enquote{Limitations on low rank approximations for covariance matrices of spatial data.}
\newblock {\em Spatial Statistics\/}, 8, 1--19.

\bibitem[\protect\citename{{U.S. Census Bureau}, }2024]{acs_data}
{U.S. Census Bureau} (2024).
\newblock \enquote{{American Community Survey Data}.}
\newblock \url{https://www.census.gov/programs-surveys/acs}.
\newblock Accessed: 2024-10-18.

\bibitem[\protect\citename{Walker and Herman, }2024]{tidycensus}
Walker, K. and Herman, M. (2024).
\newblock {\em tidycensus: Load US Census Boundary and Attribute Data as 'tidyverse' and 'sf'-Ready Data Frames\/}.
\newblock R package version 1.6.6.

\bibitem[\protect\citename{Wardrip and Pelletiere, }2008]{wardrip2008fully}
Wardrip, K.~E. and Pelletiere, D. (2008).
\newblock \enquote{Fully utilizing housing cost data in the American Community Survey PUMS Data: Identifying issues and proposing solutions.}
\newblock {\em Cityscape\/},  331--339.

\bibitem[\protect\citename{Wikle et~al., }2019]{wikle2019spatio}
Wikle, C.~K., Zammit-Mangion, A., and Cressie, N. (2019).
\newblock {\em Spatio-temporal statistics with R\/}.
\newblock Chapman and Hall/CRC.

\bibitem[\protect\citename{Zhang et~al., }2023]{zhang2023flexible}
Zhang, L., Ma, X., Wikle, C.~K., and Huser, R. (2023).
\newblock \enquote{Flexible and efficient spatial extremes emulation via variational autoencoders.}
\newblock {\em arXiv preprint arXiv:2307.08079\/}.

\end{thebibliography}
\bibliographystyle{jasa}

\end{document}